\documentclass[sigconf]{acmart}
\AtBeginDocument{%
  \providecommand\BibTeX{{%
    \normalfont B\kern-0.5em{\scshape i\kern-0.25em b}\kern-0.8em\TeX}}}

\setcopyright{acmcopyright}
\copyrightyear{2020}
\acmYear{2020}

\acmConference[KDD \textquotesingle 20 Workshop on AI for fashion supply chain]{KDD Workshop on AI for fashion supply chain}{24 August, 2020}{San Diego, USA}
\acmBooktitle{KDD \textquotesingle 20 Workshop on AI for fashion supply chain,
  24 August 2020, San Diego, USA}

\begin{document}

\title{AI Assisted Apparel Design}

\author{Alpana Dubey}
\affiliation{%
  \institution{Accenture Labs, Bangalore}
}
\email{alpana.a.dubey@accenture.com}

\author{Nitish Bhardwaj}
\affiliation{
  \institution{Accenture Labs, Bangalore}
}
\email{nitish.a.bhardwaj@accenture.com }

\author{Kumar Abhinav}
\affiliation{%
  \institution{Accenture Labs, Bangalore} 
}
\email{k.a.abhinav@accenture.com}

\author{Suma Mani Kuriakose}
\affiliation{
  \institution{Accenture Labs, Bangalore}
}
\email{suma.mani.kuriakose @accenture.com }

\author{Sakshi Jain}
\affiliation{
  \institution{Accenture Labs, Bangalore}
}
\email{sakshi.c.jain@accenture.com}

\author{Veenu Arora}
\affiliation{
  \institution{Accenture Labs, Bangalore}
}
\email{veenu.arora@accenture.com}

\renewcommand{\shortauthors}{A. Dubey et al.}
\begin{abstract}
Fashion is a fast-changing industry where designs are refreshed at large scale every season. Moreover, it faces huge challenge of unsold inventory as not all designs appeal to customers. This puts designers under significant pressure. Firstly, they need to create innumerous fresh designs. Secondly, they need to create designs that appeal to customers. Although we see advancements in approaches to help designers analyzing consumers, often such insights are too many. Creating all possible designs with those insights is time consuming. In this paper, we propose a system of AI assistants that assists designers in their design journey. The proposed system assists designers in analyzing different selling/trending attributes of apparels. We propose two design generation assistants namely Apparel-Style-Merge and Apparel-Style-Transfer. Apparel-Style-Merge generates new designs by combining high level components of apparels whereas Apparel-Style-Transfer generates multiple customization of apparels by applying different styles, colors and patterns. We compose a new dataset, named DeepAttributeStyle, with fine-grained annotation of landmarks of different apparel components such as neck, sleeve etc. The proposed system is evaluated on a user group consisting of people with and without design background. Our evaluation result demonstrates that our approach generates high quality designs that can be easily used in fabrication. Moreover, the suggested designs aid to the designer\textquotesingle s creativity. 
\end{abstract}





\begin{teaserfigure}
  \includegraphics[width=\textwidth]{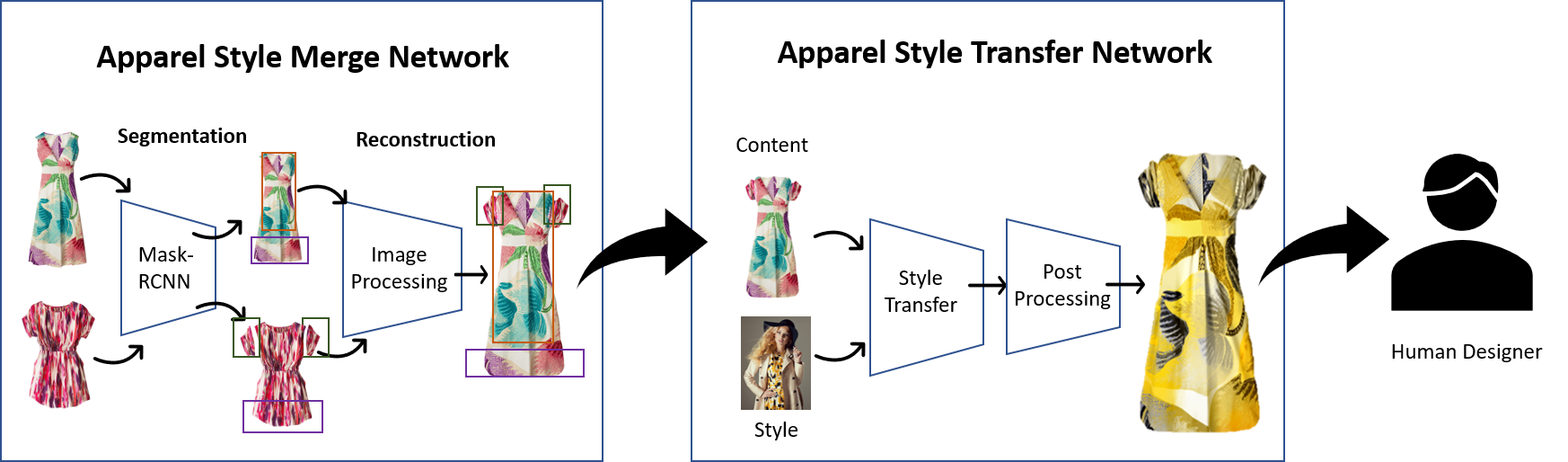}
  \caption{Schematic Flow of Creative Design Assistants Platform for Fashion (CDAP-F)}
  \Description{workflow}
  \label{fig:Workflow of CDAP-F}
\end{teaserfigure}

\maketitle

\section{Introduction}
Fashion is a fast-changing industry where designs are changed every season. The ever changing and often not well understood consumer preferences lead to huge unsold inventory which negatively impacts the business and overall margins. An estimate done by ShareCloth finds that about 30\% of apparels were never sold in 2018 \cite{Next}. Modern consumers want to have their unique identity yet remain trendy by adapting the latest fashion trends. In addition, they want instant access to new collections, ideally at price points they can afford \cite{Extent}. All these have put fashion industry in a challenging time where it needs to rapidly respond to consumers\textquotesingle  changing preferences with fresh and affordable designs at scale. At the same time, they need to reduce unsold inventory to ensure sustainability and profitability. To address this, industries are adopting new models of engagement where consumers are at the center and drive what they want through various design customization tools \cite{Next}. The need for extreme personalization has put further pressure on the designers to produce designs and their variants at a large scale.  Firstly, designers need to remain vigilant about current and upcoming trends to understand changing consumer preferences and secondly, they need to design multitude of apparels at a faster rate. Hence, there is a need of approaches to assist designers in their design process.\par
\begin{figure*}
  \centering
  \includegraphics[width=12cm,keepaspectratio]{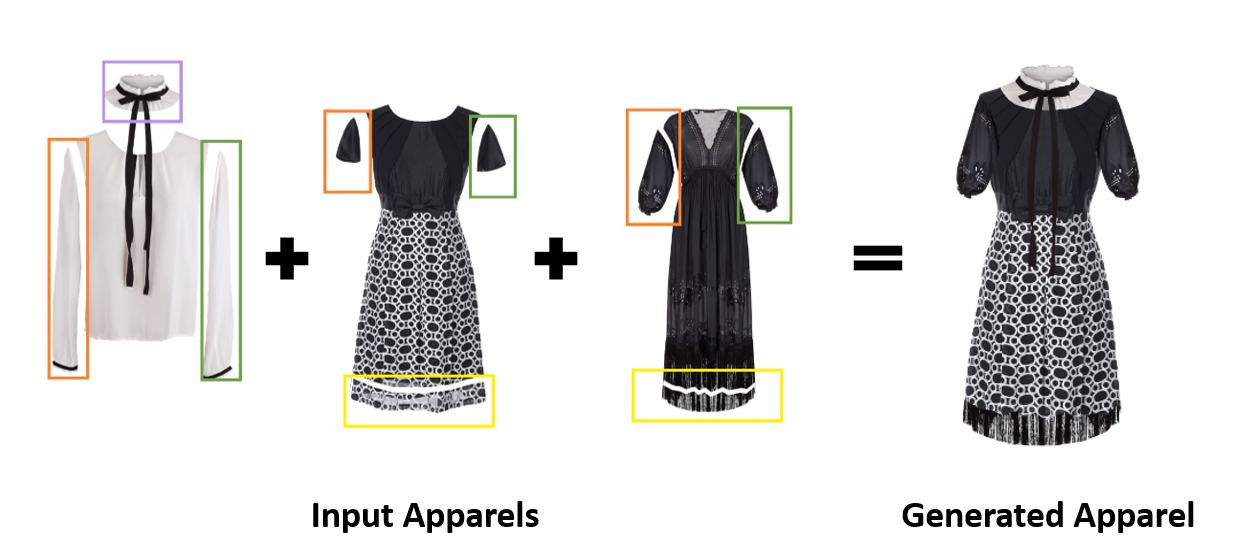}
  \caption[Our Approach]{Apparel-Style-Merge Assistant: Designer can select multiple apparels and using Apparel-Style-Merge assistant generate new design. The AI assistant segments the input apparels and generates new design by combining various segments.}
  \label{fig:Apparel-Style-Merge Assistant}
\end{figure*}

In this paper, we propose a system that assists designers in the apparel design process. With Creative Design Assistants Platform for Fashion (CDAP-F), the proposed system establishes an effective collaboration among human designers and multiple artificial intelligence assistants to complement their strengths. We propose Consumer Insights Assistant that assists designers in understanding consumer insights through sales and trends analysis. To augment designers\textquotesingle  creativity, we propose two assistants namely Apparel-Style-Merge and Apparel-Style-Transfer. Apparel-Style-Merge enables designers to combine elements from multiple apparels to create new designs (as shown in Figure \ref{fig:Workflow of CDAP-F} and Figure \ref{fig:Apparel-Style-Merge Assistant}). Apparel-Style-Transfer helps designers in customizing apparels by applying different styles, colors and patterns (as shown in Figure  \ref{fig:Workflow of CDAP-F} and Figure \ref{fig:Apparel-Style-Transfer Assistant}). We have utilized variants of deep neural networks to develop the above assistants. The proposed system has been evaluated on user group with and without design background for its usefulness. The evaluation results show that the system can significantly add creativity to designers\textquotesingle  imagination by generating good quality of unique apparel designs. Moreover, the approach being automated can produce designs at a faster rate.  \par
The main contributions of our work are:
\begin{itemize}
    \item We proposed a pipeline to generate various apparel designs using Style merge and Style transfer techniques
    \item We introduced a new dataset, DeepAttributeStyle, to serve our style merge task. The dataset contains the fine-grained segmented region of different components of apparel.
\end{itemize}
The remainder of this paper is structured as follows: Section 2 discusses the related work on apparel design. The details of the proposed method are described in section 3. We present the evaluation in section 4. Section 5 discusses threats to validity. Finally, section 6 concludes the paper with future work. 

\section{Related Work}
We present here the related work along two broad topics: first around human-AI collaboration systems in the creative field and second around AI approaches specifically meant to support apparel design. \par
Over the past decade, human and AI collaboration has evolved at a very noticeable pace. Humans now have digital colleagues and assistants to support them in their daily activities. According to a research conducted by Accenture involving 1,500 companies, it is found that firms achieve the most significant performance improvements when humans and AI work together \cite{wilson2018collaborative}. This shows that humans and AI collaboration can fetch better and productive results. There are a lot of applications being developed that involve Human-AI partnership. Such partnerships are now materializing even in creative endeavors, such as Human-AI co-design of fashion, creative writing, art generation and music composition. Stitch Fix  is a prime example of utilizing such partnership \cite{StitchFix}. Stitch Fix provides personalized shopping experience to the customer where the company picks out new clothes and sends them straight to your door, based on data you provide, such as a style survey, measurements, and a Pinterest board.  AI reduces the potential options in terms of style, size, brand, and other factors and provides a stylist with a manageable number of choices; thus, augmenting the stylist. The stylist then uses his or her expertise to finalize the package and possibly includes a personalized note. Both, human and machine are constantly learning and updating their decision making. \par

Most of the AI approaches in fashion domain are focused on apparel recommendation\cite{zhu2017your}, identifying apparel attributes\cite{s2019progressive} \cite{liu2016deepfashion}, and segmenting major apparel components\cite{liu2016deepfashion}.Apparel Design using AI has seen recent success with generative adversarial network (GAN) models and style transfer \cite{zhu2017your}\cite{ravi2019teaching} and availability of rich datasets like DeepFashion \cite{liu2016deepfashion}. \par
\textbf{GAN based apparel generation}: There has also been a growing interest in generating apparel designs using GANs, given their ability to generate appealing images. Sbai et al. \cite{sbai2018design} proposed a shape conditioned model named StyleGAN for generating fashion design images. Banerjee et al. \cite{banerjee2018let} explored different GAN architectures for context-based fashion generation. Kang et al. \cite{kang2017visually} uses Conditional GAN  to generate novel fashion items that maximize users\textquotesingle  preferences. Raffiee et al. \cite{raffiee2020garmentgan} proposed GarmentGAN to synthesize high quality images and robustly transfer photographic characteristics of clothing. The system consists of two separate GANs: a shape transfer network and an appearance transfer network. Zhu et al. \cite{zhu2017your} presented an approach for generating new clothing on a wearer based on textual descriptions. Dong et al. \cite{dong2019fashion} proposed Fashion Editing Generative Adversarial Network (FE-GAN), which enables users to manipulate the fashion image with an arbitrary sketch and a few sparse color strokes. Yu et al. \cite{yu2019personalized} proposed a personalized fashion design framework with the help of generative adversarial training that can automatically model user\textquotesingle s fashion taste and design a fashion item that is compatible to a given query item. Unlike  these approaches, we generate new designs by combining multiple apparels and then use style-transfer to add further variation. 

\begin{figure*}
  \centering
  \includegraphics[width=10cm,keepaspectratio]{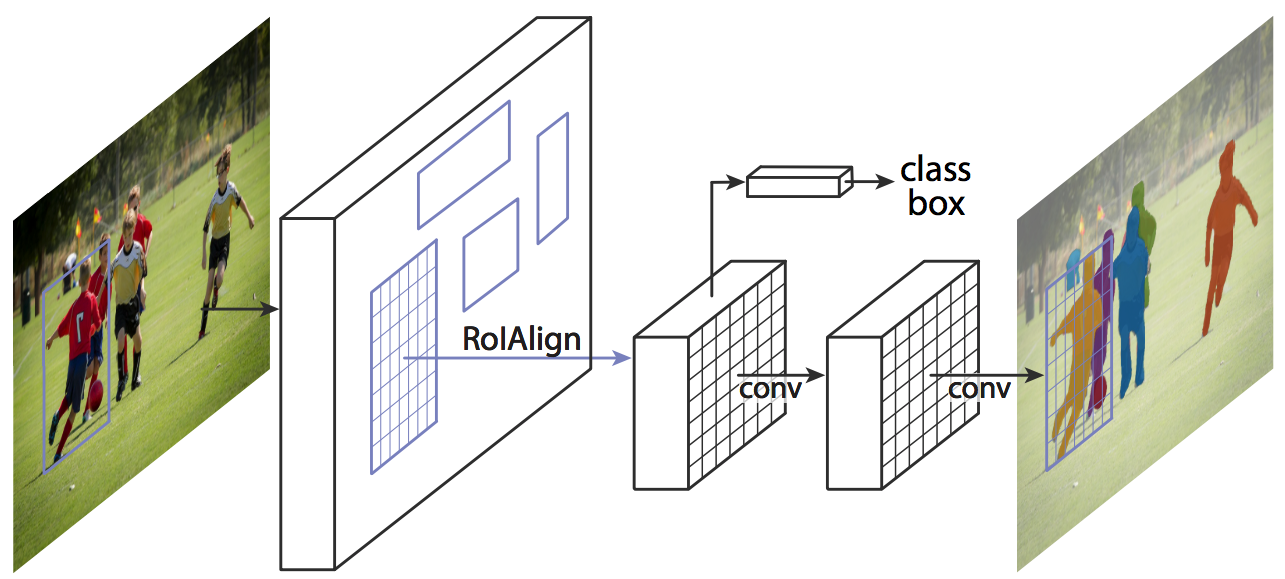}
  \caption{The Mask R-CNN framework for instance segmentation \cite{he2017mask}}
  \label{fig:Mask-RCNN Framework}
\end{figure*}

\begin{figure*}
  \centering
  \includegraphics[width=15cm,keepaspectratio]{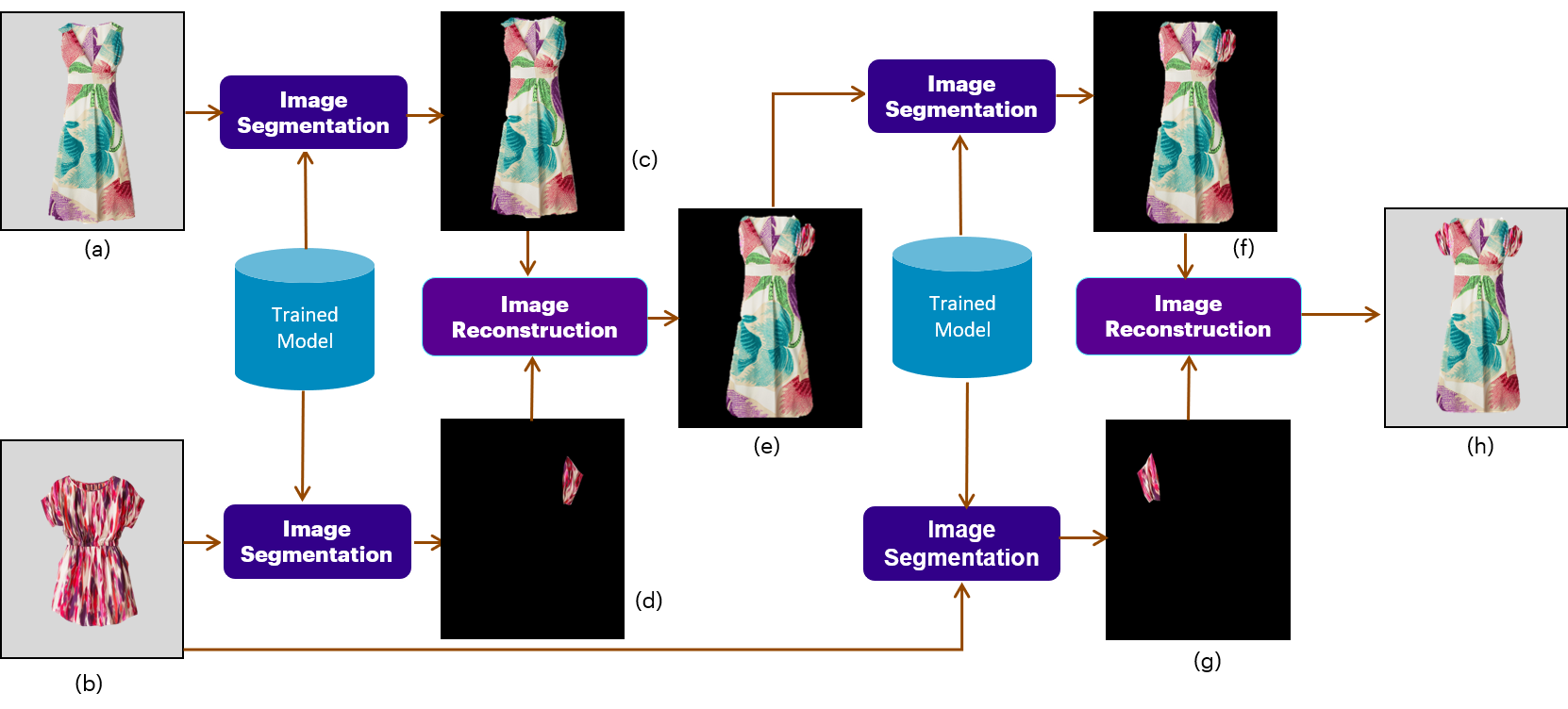}
  \caption{Apparel-Style-Merge Algorithm flow: Designer selects apparel images (a) and (b). The trained segmentation model generates different components as masks. In this example, (c) contains silhouette mask and (d) contains right sleeve mask. Image Reconstruction algorithm selects these two masks and generates intermediate output (e). (e) is taken as input for the next step and (f) is silhouette mask from (e). (g) is mask of left sleeve from (b). (f) and (g) are combined to generate the final output (h). This process is run iteratively to select multiple apparels and multiple masks to generate a lot of variations.}
  \label{fig:Apparel-Style-Merge Algorithm Flow }
\end{figure*}

\textbf{Style Transfer based apparel generation}: The seminal work of Gatys et al. \cite{gatys2016image} demonstrated the power of Convolutional Neural Networks (CNNs) in creating artistic imagery by combining the content of one image with the style of another image. Style transfer approach has been successfully applied to a wide range of applications such as social communication (e.g., Prisma \cite{Prisma}, Ostagram \cite{Ostagram}), movies, games etc. \cite{jing2019neural}. The neural style transfer algorithm has been applied to fashion to synthesize new custom clothes. Date et al. \cite{date2017fashioning} proposed an approach to personalize and generate new custom clothes using neural style transfer based on user\textquotesingle s preference and by learning the user\textquotesingle s fashion choices from a limited set of clothes from their closet. Jiang et al. \cite{jiang2017fashion} proposed a neural fashion style generator that generates a clothing image with a certain style in real-time. The global optimization stage preserves the clothing form and design and the local optimization stage preserves the detailed style pattern.  Hobley et al.\cite{hobley2018say} proposed an approach to transfer both the shape and style between images. Chen et al. \cite{chen2020tailorgan} proposed an approach to generate new fashion designs with disentangled user-defined attributes. The model generates a photorealistic image which combines the texture from reference garment image A and the new attribute from another reference image B. Style transfer with super resolution is being used to generate variety of stylized images for apparels \cite{ravi2019teaching}. These generated outputs are very similar to base design, as it uses style transfer on the base design.\par 
Overall, our approach can be differentiated from existing approaches along following lines. We have developed an end to end pipeline that involves consumer insights collection, design generation with those insights and human designer in the loop to select, filter and add more creative elements. Our approach generates high quality outputs which designers can use easily.

\section{Workflow and Technical Details }

The proposed system consists of three AI assistants hosted on Creative Design Assistants Platform for Fashion (CDAP-F): one for consumer insights and the rest two for design generation. CDAP-F enables collaboration between designers and the proposed design assistants. \par
\begin{figure*}
  \centering
  \includegraphics[width=15cm,keepaspectratio]{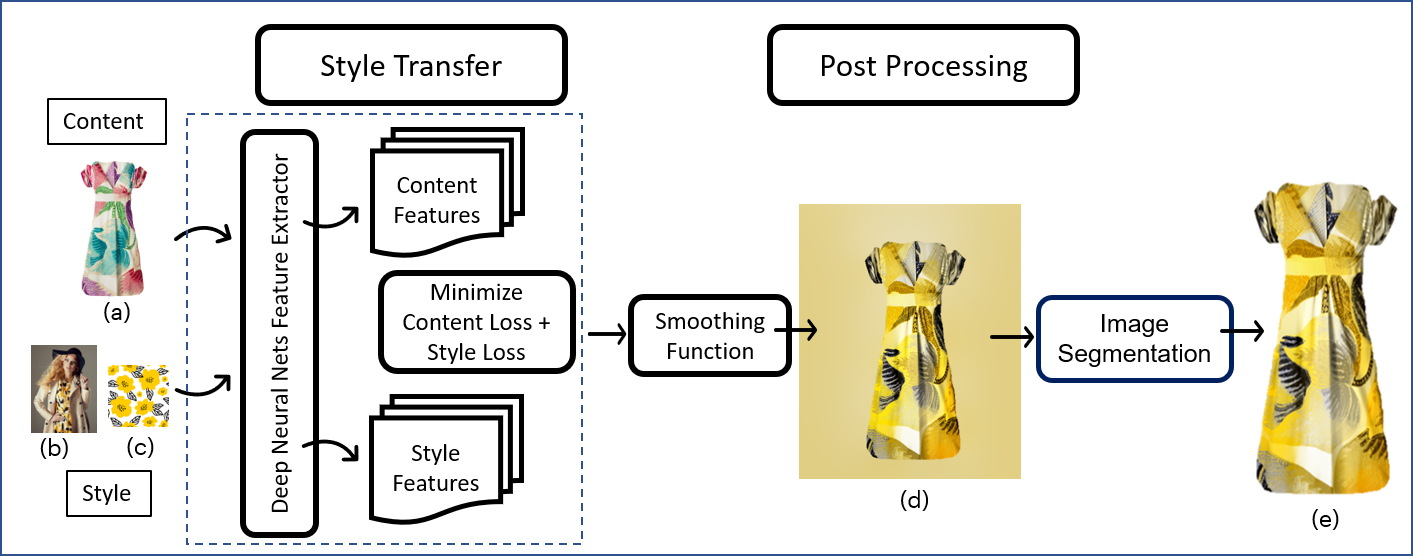}
  \caption{Apparel-Style-Transfer Algorithm flow: Designer selects generated image from Apparel-Style-Merge as content-image (a) and selects style image color pattern (c) based on a theme (b).}
  \label{fig:Apparel-Style-Transfer Assistant Workflow}
\end{figure*}
\begin{figure*}
  \centering
  \includegraphics[width=15cm,keepaspectratio]{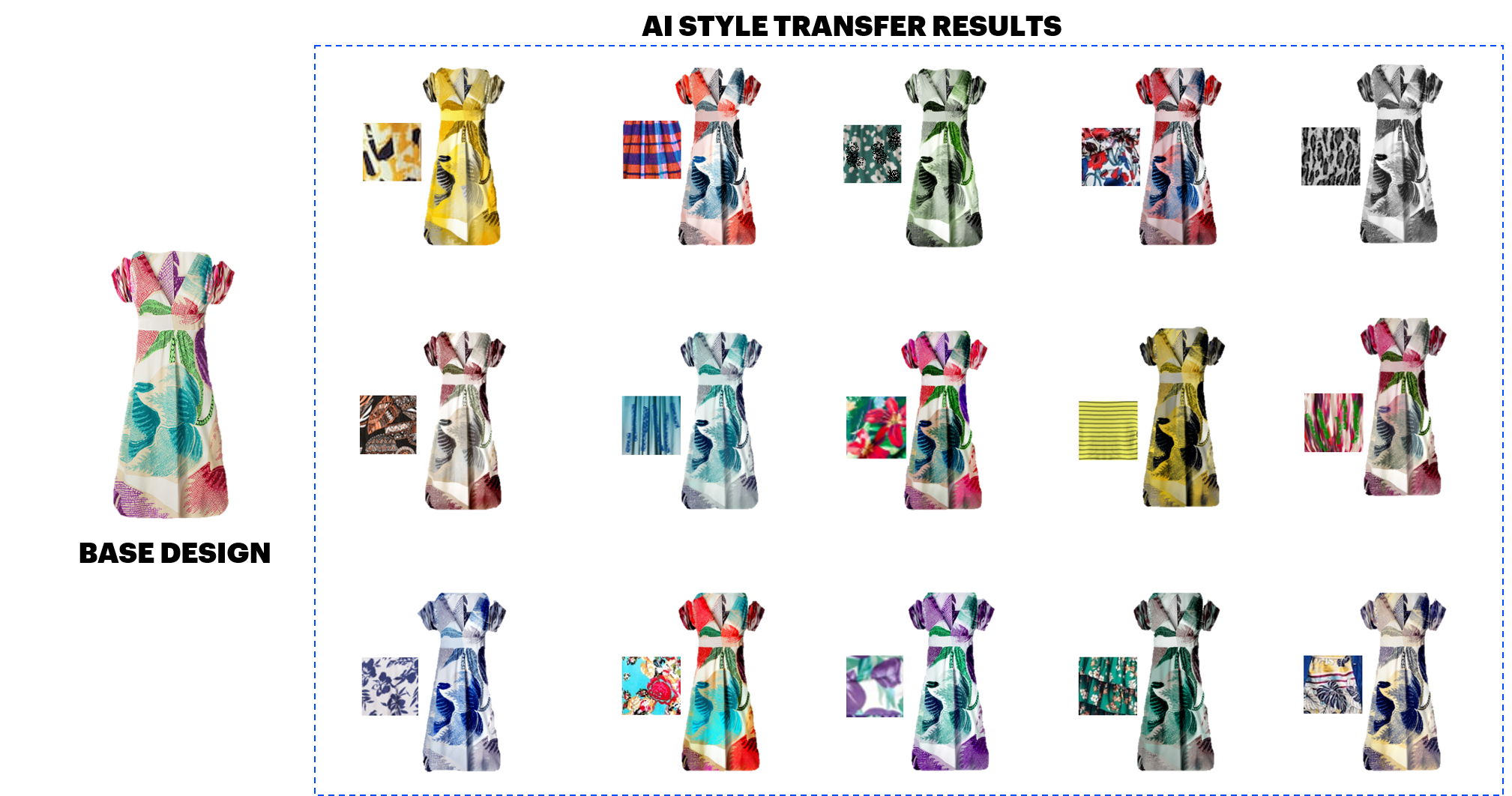}
  \caption{Apparel-Style-Transfer Assistant: Designer can select multiple trending themes/styles to draw inspirations and generate different variations}
  \label{fig:Apparel-Style-Transfer Assistant}
\end{figure*}

With the help of consumer insights assistant, designers can analyze key attributes that contribute to popularity of designs.The key attributes from any apparel can be extracted using multi-class attribute classification\cite{s2019progressive}. The popularity is identified from the sales and attributes data of designs\cite{lin2019predicting}. For instance, data may suggest that a specific color contributes significantly for sales. Designer can use such insights and select some key attributes for his new design. Human designers play an important role in collecting the requirements, understanding theme, browsing the popular designs and selecting key designs for design generation. While creating new designs, the designers may use AI assistants Apparel-Style-Merge Assistant (as shown in Figure \ref{fig:Apparel-Style-Merge Assistant} and Figure \ref{fig:Apparel-Style-Merge Algorithm Flow }). Further, designer may generate different variations of a design using Apparel-Style-Transfer Assistant (as shown in Figure \ref{fig:Apparel-Style-Transfer Assistant Workflow} and Figure \ref{fig:Apparel-Style-Transfer Assistant}). In the next subsections, we will talk about technical details of AI assistants.\par

\subsection{Apparel-Style-Merge Assistant}
Apparel-Style-Merge assistant works on two high level steps: 1. segmentation of input designs, and 2. reconstruction of new designs by placing segmented parts from multiple apparels at the appropriate places. \par

Fashion Design using AI has seen recent success with Deep Neural Network based generative models \cite{zhu2017your}\cite{ravi2019teaching} and availability of detailed dataset like DeepFashion \cite{liu2016deepfashion}. However, existing datasets are not sufficiently enriched to be meaningfully used in other applications. For instance, DeepFashion Dataset consists of apparel images, categories, and high-level regions like top, bottom and full dress but it lacks detailed regions like silhouette, sleeve, collar, shoulder, etc. To develop our system, we created a new dataset, DeepAttributeStyle, which contains regions with more details. DeepAttributeStyle is annotated with rich information of apparels. We hired ten expert designers who manually created masks for major segments of the apparels such as: 0-BackGround, 1- Silhouette, 2-Collar, 3-Neck, 4-Print, 5-Hemline, 6-Sleeve-right, 7-Sleeve-left, 8-Shoulder-right and 9-Shoulder-left. For tagging images with these regions, we used publicly available tagging tool, VIA (VGG Image annotator) \cite{dutta2019vgg} \cite{dutta2016via}. The VIA software allows human annotators to define and describe regions in an image. The manually defined regions can have one of the following six shapes: rectangle, circle, ellipse, polygon, point and polyline. We used Polygon shaped regions that captures the boundary of objects having a complex shape. We demonstrated the tool usage to the expert designers for tagging the apparels. \par

Image styling and segmentation are the major building blocks of many of the AI approaches that deal with designs. Recent advancements in AI have resulted in great improvement in image segmentation and image reconstruction. For solving the problem of Image segmentation (as shown in Figure \ref{fig:Mask-RCNN Framework}), the latest Deep Neural Network based models like RCNN, Faster-RCNN and Mask-RCNN \cite{he2017mask} have produced results with great accuracy. From the experimental results, Mask-RCNN has outperformed other state-of-the-art solutions like Faster-RCNN, InstanceCut, DWT, etc. Therefore, we used Mask-RCNN framework for our segmentation model. We removed the final Softmax layer of Mask RCNN segmentation model and retrained the model on our dataset, DeepAttributeStyle.\par

We used a dataset of ~500 apparel images to conduct our experiments.  We tagged ~500 images with the defined masks. The dataset is trained using state-of-the-art Mask RCNN model for segmentation of these regions. For this purpose, we changed the number of outputs in SoftMax layer with our number of segments, i.e., 10.  We split our dataset into train(80\%), validation(10\%) and test(10\%) set. We used 400 images for training, 50 images for validation and 50 images for testing.\par
For evaluation of segmentation module, we used segmentation model on test dataset and calculated IOU(Intersection-of-union) score  for each class. IoU metric computes the number of pixels overlapping between the target and prediction masks divided by the total number of pixels present across both masks. The IOU score shows good accuracy for classes like silhouette, hemline and sleeves (as shown in Table 1).

\begin{equation}
IoU score =  \frac{target \cap prediction}{target \cup prediction}
\end{equation}

\begin{table}[h!]
  \centering
  \caption{IOU score for Segmentation Model }
  \begin{tabular}{ |p{4cm}|p{3cm}|  }
    \hline
    \bfseries Attribute name& \bfseries IoU score\\
    \hline
  Silhouette   & 0.90\\
  \hline
  Hemline   & 0.81\\
    \hline
    Sleeve-right   & 0.78\\
      \hline
    Sleeve-left   & 0.76\\
    \hline
    Neck   & 0.57\\
    \hline
    Shoulder-right   & 0.57\\
    \hline
    Shoulder-left   & 0.55\\
    \hline
    Print   & 0.48\\
    \hline
  \end{tabular}
  \label{table:2-IoU score}
\end{table}

We used the trained segmentation model for segmentation of different components of apparels and created new apparel design by using image reconstruction algorithm  by combining different segments from different apparels. The image reconstruction algorithm (as shown in Figure \ref{fig:Apparel-Style-Merge Algorithm Flow }) takes two inputs, which are masks from two apparels. We generate new apparel using bitwise addition of these masks with original images. The algorithm is run multiple times to combine different masks and generate multiple new designs. The AI assistant identifies elements of existing dresses and understands their positioning and then automatically creates variations leading to novel designs.

\begin{figure*}
  \centering
  \includegraphics[width=15cm,keepaspectratio]{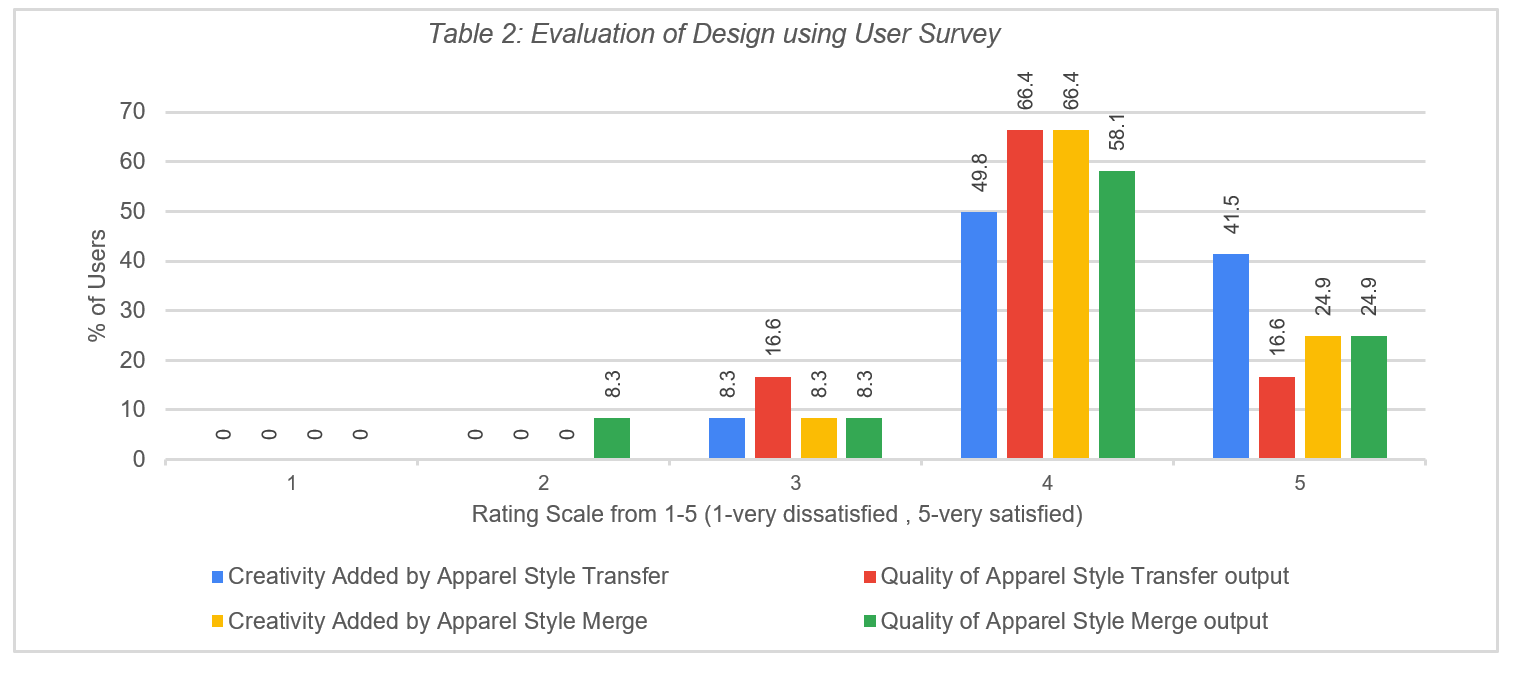}
  \caption*{}
  \label{fig:Evaluation from user survey}
\end{figure*}

\subsection{Apparel-Style-Transfer Assistant}
Style Transfer has been used for creating new art forms in various industries. We propose a novel approach for Style Transfer of apparel designs. We extended the style transfer approach for photo-realistic stylization \cite{huang2017arbitrary} \cite{li2018closed} and then used semantic segmentation \cite{he2017mask} for generating better quality of outputs. The solution takes two inputs; viz; content image, which is coming from Apparel-Style-Merge Assistant and style image which is inspired from latest theme or trends. Our solution pipeline (as shown in Figure \ref{fig:Apparel-Style-Transfer Assistant Workflow}) consists of following steps: 1. Applying style transfer on the whole image and 2. Leveraging DeepAttributeStyle segmentation model to crop only the dress. The segmentation model is further used by Apparel-Style-Transfer to segment silhouette from the stylized image to make the stylized image more photorealistic. This pipeline has resulted in faster-photorealistic style transfer. \par
In our solution, we use the concept of style transfer to generate and visualize innumerous photorealistic designs at scale (as shown in Figure \ref{fig:Apparel-Style-Transfer Assistant}) in very short time, ~1 minute per design, thus augmenting designers\textquotesingle  creativity and increasing design efficiency. The generated designs in our solution are different from base designs as we use multiple apparels as input and then use style image to further add variation (as shown in Figure \ref{fig:Workflow of CDAP-F}).

\section{Evaluation}
We conducted following evaluations to assess our system: 1. Evaluation of our assistants for fast fashion designs, 2. Quality of generated designs, 3. Usefulness of generated designs in creativity augmentation. 
\par In order to understand the design process and validate our solution, we conducted study with professors from a Fashion University. From this study, we observed that the design process involves the selection of a theme/mood, color palette and fabric. The predominant tools used by them are CoralDraw \cite{CoralDraw} and Adobe Photoshop \cite{Photoshop}. To evaluate the platform along design time, we have compared end-to-end time taken from selection of two apparels and creation of new apparel  with and without CDAP-F. Without CDAP-F, it takes 5 minutes to 40 minutes with an average time of 10-15 minutes whereas with CDAP-F it takes 2-3 minutes.  
\par For evaluation of quality and creativity augmentation, we conducted a survey to evaluate end-to-end CDAP-F. We took feedback from 15 designers for a set of 9 questions. The survey participants were fashion designers with 1 to 6 years of designing experience. The participants consisted of 9 male and 6 female designers. The questions were framed to evaluate the quality of generated outputs and to rate the creativity added by the AI assistants. The responses were recorded on a scale of 1-5 (1-very dissatisfied, 5-very satisfied). \par 
The questions for evaluation of Apparel-Style-Merge Assistant were as follows: 1) How realistic are the generated designs? 2) How different are the generated images to the base images? 3) Does generated design need re-sketching before giving for sample creation? 4) Do generated images assist in designers\textquotesingle  imagination? The questions for Apparel-Style-Transfer Assistant were : 1) How realistic are the styled designs? 2) How close are the styled images to the base images? 3) Rate the color distribution of the styled images. 4) Does it aid to your imagination? 5) Does design need re-designing before giving for sample creation?
\par Based on the responses from the designers (as shown in Table 2), we concluded that the tool helped the designers to get inspired from a lot of design ideas. About 90\% users (who rated 4 and 5) agreed that CDAP-F added to designers\textquotesingle  creativity. More than 80\% users (who rated 4 and 5) were satisfied with the quality of generated apparels. One of the survey participants gave feedback that the tool greatly helped her to derive new inspirations and ideas in short time. CDAP-F is a truly unique Human-AI collaborative tool which helps to bridge the gap between human creativity and machine efficiency. Our tool generated multiple high quality outputs which help the designers to visualize and envision the final product.

\section{Threats to Validity}

Our results show great promises in CDAP-F.  However, there are some threats to validity to these studies as follows. Firstly, the experiments are conducted on limited number of subjects  due to unavailability of human subjects at scale. Secondly, the assessment done by individuals on survey questions can be subjective. However, as we see a large concurrence on the ratings which was further confirmed by senior professors from Department of Fashion, IIS University, we believe these results would hold true if conducted on larger userbase.

\section{Conclusion and Future Work}
We proposed a novel approach for apparel design where AI augments human designers in multiple ways. With our system, designers can derive inspiration from existing designs and automatically generate high-quality designs. With significant reduction in design cost and faster time to market, the proposed approach can transform not just the fashion industry but also other similar industries where product\textquotesingle s form and shape play an important role. We believe that the approach can be further extended with other approaches to improve its usefulness. For instance, Generative Adversarial Network (GAN) based models \cite{zhu2017your} with Super-Resolution models can be used to generate randomized and high-quality images \cite{ravi2019teaching}. Image generation models guided by analytical engine can be used to predict salability of the generated designs which can help designers in shortlisting designs. We plan to extend our approach with these capabilities in future version of CDAP-F.

\begin{acks}
To Dr. Sunetra Datt, Sr. Assistant Professor and Ms. Vidushi Vashishtha, Assistant Professor at IIS University, Department of Fashion and Textile, Jaipur, India for explaining design process and discussion on our solution.
To Dhruv Bajpai, Sr. Manager, Accenture, India for insights into business aspects for designing apparels.
\end{acks}

\bibliographystyle{ACM-Reference-Format}
\bibliography{sample-base}


\end{document}